%% file: main.tex
\documentclass{article}

\usepackage[preprint]{neurips_2024}

\usepackage{times}
\usepackage{soul}
\usepackage{graphicx}
\usepackage{caption}
\usepackage{subcaption}
\usepackage{amsmath}
\usepackage{amsthm}
\usepackage{booktabs}
\usepackage{makecell}
\usepackage{algorithm}
\usepackage{algorithmic}
\usepackage{pgfplots}
\usepackage[utf8]{inputenc} 
\usepackage[T1]{fontenc}    
\usepackage{hyperref}       
\usepackage{url}            
\usepackage{booktabs}       
\usepackage{amsfonts}       
\usepackage{nicefrac}       
\usepackage{microtype}      
\usepackage{xcolor}         

\usepackage{xcolor}
\usepackage{xspace}
\usepackage{wrapfig}

\newcommand{\ES}{\textcolor{teal}{\textbf{EvS}}\xspace}

\title{MLEM: Generative and Contrastive Learning as Distinct Modalities for Event Sequences}

\author{
  Viktor Moskvoretskii$^*$ \\
  Skolkovo Institute of Science and Technology\\
  HSE University\\
  \texttt{V.Moskvoretskii@skoltech.ru} \\
  \And
  Dmitry Osin$^*$ \\
  Skolkovo Institute of Science and Technology\\
  \texttt{d.osin@skoltech.ru} \\
  \And
  Egor Shvetsov \\
  Skolkovo Institute of Science and Technology\\
  \texttt{e.shvetsov@skoltech.ru} \\
  \And
  Igor Udovichenko \\
  Skolkovo Institute of Science and Technology\\
  \texttt{i.udovichenko@skoltech.ru} \\
  \And
  Maxim Zhelnin \\
  Skolkovo Institute of Science and Technology\\
  \texttt{M.Zhelnin@skoltech.ru} \\
  \And
  Andrey Dukhovny \\
  Sberbank\\
  \texttt{AADukhovny@sberbank.ru} \\
  \And
  Anna Zhimerikina \\
  Sberbank\\
  \texttt{Zhimerikina.A.Y@sberbank.ru} \\
  \And
  Evgeny Burnaev \\
  Skolkovo Institute of Science and Technology\\
  AIRI\\
  \texttt{e.burnaev@skoltech.ru} \\
}

\begin{document}

\maketitle

\begin{abstract}


    
    
    
    


This study explores the application of self-supervised learning techniques for event sequences. It is a key modality in various applications such as banking, e-commerce, and healthcare. However, there is limited research on self-supervised learning for event sequences, and methods from other domains like images, texts, and speech may not easily transfer. To determine the most suitable approach, we conduct a detailed comparative analysis of previously identified best-performing methods. We find that neither the contrastive nor generative method is superior. Our assessment includes classifying event sequences, predicting the next event, and evaluating embedding quality. These results further highlight the potential benefits of combining both methods. Given the lack of research on hybrid models in this domain, we initially adapt the baseline model from another domain. However, upon observing its underperformance, we develop a novel method called the Multimodal-Learning Event Model (MLEM). MLEM treats contrastive learning and generative modeling as distinct yet complementary modalities, aligning their embeddings. The results of our study demonstrate that combining contrastive and generative approaches into one procedure with MLEM achieves superior performance across multiple metrics. 


\end{abstract}

\def\thefootnote{*}\footnotetext{These authors contributed equally to this work}

\input{content/introduction}

\input{content/preliminary}

\begin{figure*}[t] 
  \centering
  \includegraphics[width=1\textwidth]{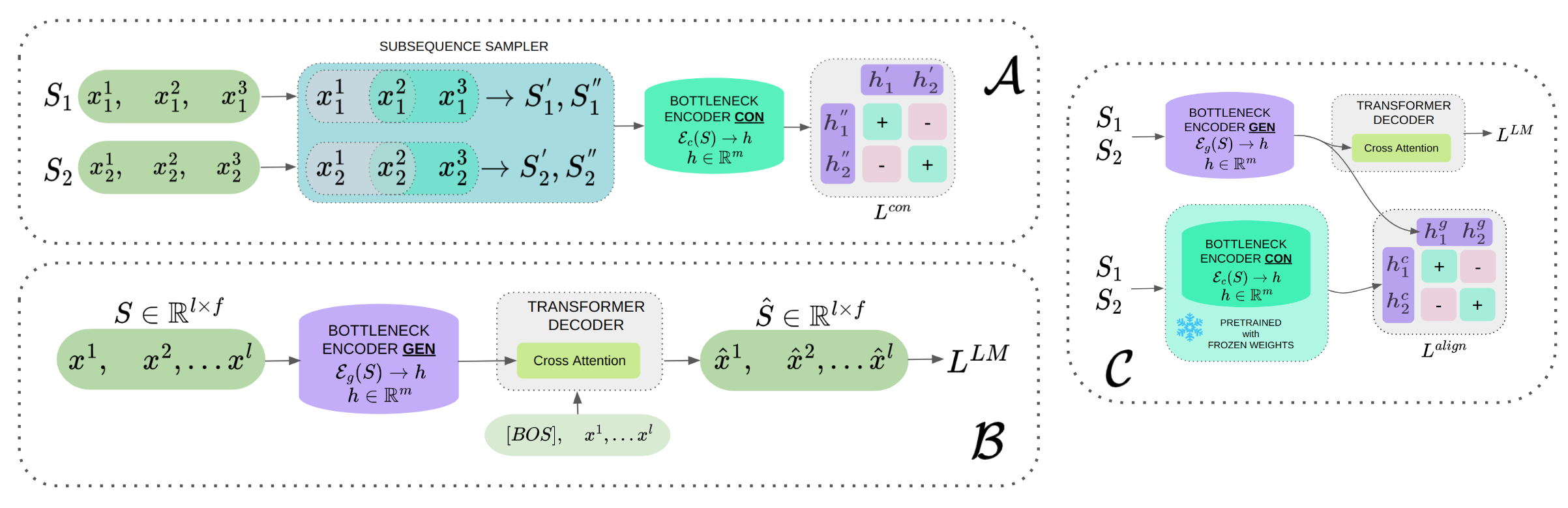}
\caption{
    \textbf{Schematic Models Overview:} \textbf{A)} The Contrastive Model divides sequence $S_i$ into subsequences $\{S_i', S_i''\}$, encodes them into latent vectors $\{h_i', h_i''\}$, and applies contrastive loss $L_{\text{con}}$. \textbf{B)} The Generative Model autoregressively reconstructs sequence $S$ cross-attended to latent vector $h$, derived from $S$ via a bottleneck encoder. Language model loss $L_{\text{LM}}$ is applied to reconstructed sequence $\hat{S}$. \textbf{C)} The MLEM Hybrid Model integrates generative and contrastive approaches, treating them as distinct modalities. It generates latent vectors $h_i^g$ and $h_i^c$ from a learned generative encoder and frozen, pretrained contrastive encoder respectively. An alignment loss $L_{\text{align}}$ is utilized to align vectors from the same sequence while separating those from different sequences. Language modeling loss $L_{\text{LM}}$ is applied as in generative model.
  }
  \label{fig:contrastive}
\end{figure*}

\section{Methodology}


\subsection{Contrastive Learning}  \label{Methodology:contrastive}

For our contrastive learning approach, we employ COLES~\citet{babaev2022coles} framework, as it has shown superior performance over other contrastive methods. It learns to encode the sequence $S$ into a latent vector $h \in  \mathcal{R}^m$ by bringing sub-sequences of the same sequence closer in the embedding space, while pushing sub-sequences from different sequences further apart. The overall procedure is illustrated in Figure~\ref{fig:contrastive} and formal definition is given in Appendix~\ref{appendix:COLES}

\subsection{Generative Modeling} \label{Methodology:Generative}

For our generative model, we consider the best performing setting from GNTPP~\citep{lin2022exploring} and adapt it to our task.

For the encoder, we utilize RNN, as it shows dominance to our task in recent studies~\citep{Udovichenko_2024}. It extracts the final hidden state $h$, which is subsequently passed into a Transformer decoder \citep{vaswani2023attention}. The decoder autoregressively reconstructs the entire sequence, conditioned on $h$ via the Cross-Attention mechanism. This entire process is illustrated in Figure~\ref{fig:contrastive}.

To train the model, we use the next event prediction objective, which is similar to training language models.  Nonetheless, as each event in the sequence $x^j_i=(t^j_i,\Upsilon^j_i)$ contain multiple sub-events $\Upsilon=\{k^1,k^2, \ldots\}$, we need to reconstruct all of them. To this end, we apply the cross-entropy loss for each categorical $k$ and the Mean Squared Error (MSE) loss for each real-valued $k$. The final loss is a cumulative sum of the losses for all elements in $\Upsilon=\{k^1,k^2, \ldots\}$, plus the MSE loss for time $t$. We denote this loss as $L^{LM}$.

Intuitively, this procedure requires our encoder to develop a representation informative enough for the decoder to accurately reconstruct the entire sequence. Since we map all our sequences  $S$ into, $h \in \mathcal{R}^m$ we call all encoders bottleneck encoders.


\input{content/fine_tune_table}


\subsection{Na\"ive Hybrid Method}
We consider weighted summation of contrastive and generative loss as baseline hybrid model, as it showed dominant performance in previous studies~\citep{kim2021hybrid}. We call this approach Na\"ive since it is a straightforward combination of losses $L=\alpha L^{LM}+\beta L^{con}$.


\subsection{MLEM}

In our method, named MLEM, we propose to combine contrastive learning and generative modeling by treating them as distinct modalities with further alignment.
To align the embeddings obtained through generative modeling with those acquired through contrastive learning, we employ a contrastive aligning procedure inspired by CLIP~\citet{radford2021learning}. We utilize the recent SigLIP loss~\citet{zhai2023sigmoid} for this purpose. 

Overall \textbf{MLEM} training procedure is nearly the same as for the generative model described in Section~\ref{Methodology:Generative}, except the alignment, which incurs an additional loss~(\ref{eq:align}) resulting in the total loss~(\ref{eq:full_loss}). Unlike CLIP and other multimodal models, our approach involves pre-training only one modality. Specifically, we pre-train the contrastive encoder first and then train the generative encoder, aligning it with the fixed contrastive encoder.

Formally, both generative $\mathcal{E}_{g}$ and contrastive $\mathcal{E}_{c}$ encoders receive a set of sequences $C=\{S_1, S_2, \ldots\}$ and map each sequence $S$ into corresponding latent vectors $h^g$ and $h^c$. The goal of \textbf{MLEM} alignment is to pull embeddings obtained from the same sequence closer to each other and to push embeddings from different sequences further away.  This procedure is illustrated in Figure~\ref{fig:contrastive}. Below is the alignment loss we use:


\begin{equation} \label{eq:align}
L^{align} = \frac{1}{|C|} \sum_{i=1}^{|C|} \sum_{j=1}^{|C|} \underbrace{\log \frac{1}{1 + e^{z_{ij} (-t\cdot {h^g_i} \cdot h^c_j+b)}}}_{L^{align}_{ij}}
\end{equation}

Here,  \( z_{ij} \in \{-1, 1\}\) indicates if a pair of embeddings originated from the same sequence $S$, \( t \) and \( b \) denote the temperature and bias, respectively, both are learnable parameters.

\medskip

Total loss to train \textbf{MLEM}:
\begin{equation} \label{eq:full_loss}
L = \alpha L^{LM}(S, \hat{S}) + \beta L^{align}(\mathcal{E}_{g}(S), \mathcal{E}_{c}(S))
\end{equation}

Here, $S$ and $\hat{S}$ denote original and reconstructed sequences, $\alpha$ and $\beta$ are hyperparameters that adjust the strength of alignment. 



\section{Experiments} \label{Experiments}

\subsection{Datasets}

In our study, we use five established \ES datasets formally outlined in Section~\ref{sec:prelimenaries}. 
\begin{itemize}
    \item A subset of \ES datasets from SeqNAS benchmark~\citet{Udovichenko_2024}, more precisely: ABank, Age, TaoBao. ABank and Age are bank transactions, and TaoBao is user activity. 
    \item PhysioNet 2012 comprises highly sparse medical event data, with the goal of predicting mortality \citet{physionet}.  
    \item Pendulum is a synthetic dataset simulating pendulum motion with the objective of predicting its length using a sequence of coordinates. This dataset was created specifically to evaluate a model's capability to effectively incorporate the time component in sequence modeling.
\end{itemize}
The detailed description is provided in Appendix~\ref{appendix:dataset}.

\input{content/probing_table}

\subsection{Evaluation objectives} \label{sec:objectives}

To evaluate the effectiveness of self-supervised training strategies, we follow previous studies \citet{dubois2024evaluating,dubois2021learning} and focus on two key aspects: 
\begin{itemize}
    \item Fine-Tuning Performance. The downstream metrics after fine-tuning the entire network.
    \item Embeddings Probing. The quality of the embeddings for downstream and TPP tasks tested via linear and non-linear probing.
\end{itemize} 

As well, we include additional objectives for embedding quality, that were discovered in previous studies \citet{razzhigaev2023shape,nakada2020adaptive} , that are connected with performance:
\begin{itemize}
    \item Anisotropy. It assesses the non-uniformity of embeddings in space, providing insights into the contextualization of our embedding. Lower anisotropy in embeddings has been linked to better model performance \citet{ait-saada-nadif-2023-anisotropy}.
    \item Intrinsic dimension. It evaluates the optimal dimensionality of data, showing information captured by the embeddings. 
\end{itemize}

Objectives details are described in appendix~\ref{appendix:objectives}.

\subsection{Models} \label{Experiments:Models}

\textbf{Base model and Feature embeddings}. To transform features into a numerical vector, we use an embedding layer for categorical features and linear projection for numerical ones. For ABank, Age and TaoBao we set the dimension of each feature to 32. For Pendulum and PhysioNet dimension is set to 8. Additionally, we use the time difference between events as a separate feature.

\smallskip
\textbf{We use GRU} as an encoder because it has been proven effective in encoding time-ordered sequences ~\citet{rubanova2019latent,tonekaboni2021unsupervised,NEURIPS2019_c9efe5f2,Udovichenko_2024}. \citeauthor{Udovichenko_2024} performs a Neural Architecture Search on various models and highlights that the presence of RNN blocks in the architecture is associated with higher performance on average. \citeauthor{lin2021empirical} illustrate that the performance gain brought by history encoders is very small for TPP modeling. Based on these empirical observations, we believe that our design choice is reasonable.

We it with a single-layer and a hidden size of 512, we take the last hidden state as sequence embedding. For consistency in our experiments, we use the same hyperparameters across all training strategies and datasets.

\smallskip

For \textbf{Supervised} model, we use the aforementioned encoder with a linear head. 

\medskip

\textbf{Contrastive} learning approach is described at~\ref{Methodology:contrastive} and uses the same encoder as mentioned above. Furthermore, we integrate a Feed Forward Projector atop the GRU for loss calculation, a technique commonly adopted in several studies for enhanced performance~\citet{grill2020bootstrap,oquab2023dinov2}.

\medskip

For the \textbf{generative} modeling, we employ a vanilla transformer decoder configured with LayerNorm, consisting of 3 layers, 2 heads, a hidden size of 128, and a Feed Forward hidden size of 256. To ensure training stability, we also apply LayerNorm to the sequence embedding produced by the encoder. Additionally, a projection layer is used atop the decoder to predict each feature.

\subsection{Training details}

To maintain consistency in our experiment, we trained all models using identical parameters. Models were trained for 100 epochs on datasets with fewer than 100K sequences and for 40 epochs on larger datasets. The learning rate was set at 1e-3, weight decay at 3e-3, and the batch size was fixed at 128. For the Hybrid models, we set 
$\alpha$ at 1 and $\beta$ at 10. We averaged the results across three different seeds to ensure reliability and reproducibility. The standard deviation could be not reported if negligible.

\section{Results}

\subsection{Fine-tuning performance}

First, we evaluate the effectiveness of fine-tuning the entire model after pre-training procedure. The results presented in Table~\ref{tab:fine-tune} indicate that, among all the evaluated pre-training methods, \textbf{MLEM} consistently achieves the most favorable results across all datasets, except for the highly sparse PhysioNet dataset. Furthermore, it is noteworthy that solely supervised learning consistently produces inferior results.

\subsection{Embeddings probing}

We conducted a comprehensive analysis of the quality of embeddings using both linear and non-linear probing techniques, results are presented in Table~\ref{tab:linear-non-linear}. 

Our evaluation reveals that neither the contrastive nor the generative approach consistently outperforms the other. However, by effectively integrating the advantages of both contrastive and generative techniques, \textbf{MLEM} model demonstrates superior performance across most datasets on average. Notably, when \textbf{MLEM} is not the top performer, it is often ranked second, highlighting its versatility.

\begin{table}[t]
\centering
\begin{minipage}{0.43\textwidth}
\caption{Average anisotropy and intrinsic dimension across datasets}
\centering
\resizebox{\textwidth}{!}{%
\begin{tabular}{lcc}
\hline
Method      & Anisotropy $\downarrow$ & Intrinsic Dimension $\uparrow$ \\ \hline
Contrastive & $0.11 \pm 0.04$  & $10.15 \pm 6.12$           \\
Generative  & $0.08 \pm 0.03$  & \underline{$14.86 \pm 10.12$}            \\
Na\"ive     & \underline{$0.07 \pm 0.04$}  & $11.26 \pm 6.51$             \\
MLEM        & $\mathbf{0.06 \pm 0.03}$  & $\mathbf{15.86 \pm 9.02}$             \\ \hline
\end{tabular}
}
\label{tab:small_AID}
\end{minipage}%
\hspace{3mm} 
\begin{minipage}{0.53\textwidth}
\caption{Performance change in linear probing performance after shuffling events order, higher is better}
\centering
\resizebox{\textwidth}{!}{%
\begin{tabular}{lccccc}
\toprule
Method & ABank & Age & PhysioNet & Pendulum & TaoBao \\
\midrule
Contrastive & +0.47\% & -1.93\% & -0.37\% & -367.57\% & -0.15\% \\
Generative & -0.93\% & +0.16\% & -6.19\% & -91.89\% & -0.29\% \\
Naive & -1.42\% & +0.99\% & -3.00\% & -147.46\% & 0.00\% \\
MLEM & -1.45\% & -0.16\% & -6.56\% & -285.37\% & +0.29\% \\
\bottomrule
\end{tabular}
}
\label{tab:performance_shuffle}
\end{minipage}
\end{table}

\subsection{Anisotropy and Intrinsic Dimension} \label{sec: anisotropy}

We conducted an evaluation of anisotropy and the intrinsic dimensions of embeddings from various pre-training strategies, as detailed in \ref{sec:objectives}. 

Results presented in Table~\ref{tab:small_AID} show that \textbf{MLEM} has the highest intrinsic dimension and the lowest anisotropy, on average, across all datasets. This finding may indicate that our pre-training strategy effectively enables the embeddings to encapsulate a greater amount of information and ensures their uniform distribution in space. 


However, we found no correlation between intrinsic dimension or anisotropy and downstream performance in our domain, as illustrated in correlation plot~\ref{fig:corr_aid}. While our findings align with the most recent study by \citet{dubois2024evaluating} regarding anisotropy, they differ in that they observed a correlation between intrinsic dimension and performance.

\begin{figure}[t!] 
  \centering
  \includegraphics[width=0.49\textwidth]{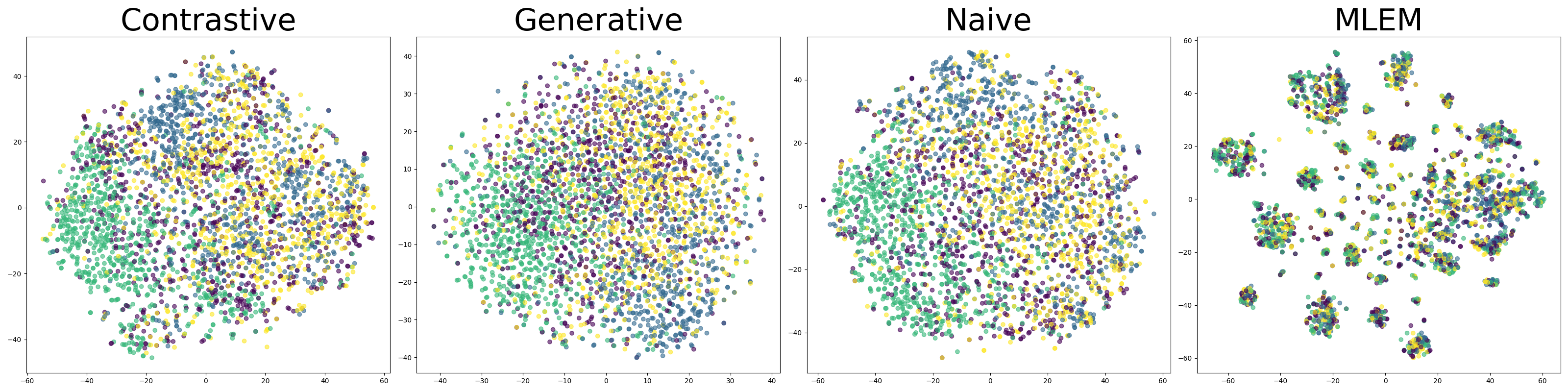}
    \includegraphics[width=0.49\textwidth]{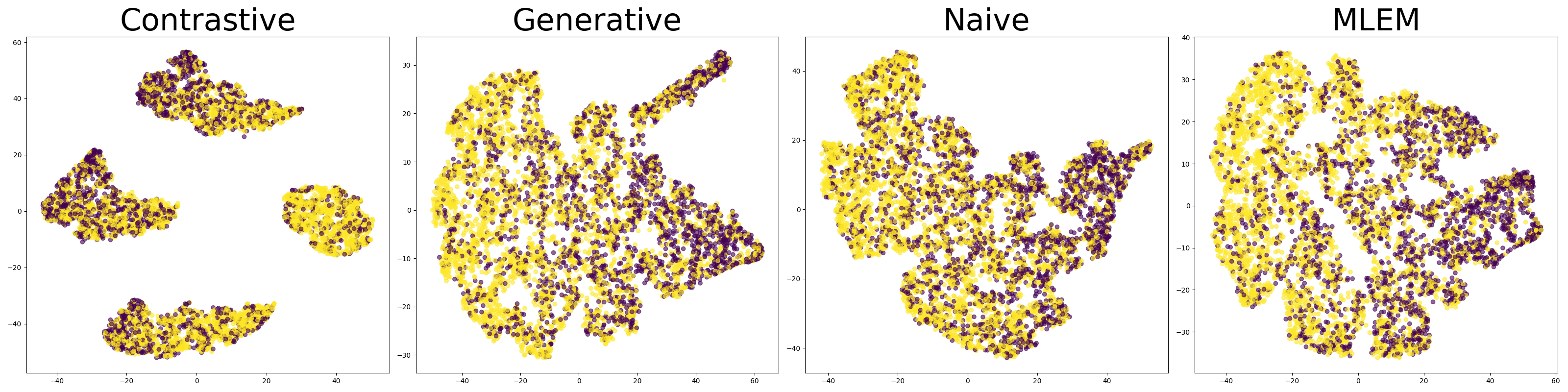}

  \caption{The t-SNE visualizations showcase embeddings resulting from various pre-training strategies. The left portion displays the embeddings for the Age dataset, while the bottom row illustrates those for the TaoBao dataset. Each point represents a sequence $S_i$ from a given dataset, colored accordingly to the corresponding attribute $y_i$. For Age, there are 4 classes and 2 classes for TaoBao}
  \label{fig:tsne}
\end{figure}

\subsection{Visualization} \label{sec:visual}


The findings related to anisotropy and intrinsic dimension align with the t-SNE visualization of the Age and TaoBao datasets shown in Figure~\ref{fig:tsne}. While contrastive embeddings typically form distinct clusters, the generative approaches and MLEM provide a more complex embedding space. 

\section{Embedding Robustness}

Our study also explores the robustness of model embeddings to specific types of perturbations. We focus on examining the impact of dropout and the shuffling of events within one sequence on the linear probing performance.

We assess robustness with classification or regression tasks. 
In each scenario, we assess either the class of the whole sequence or the real value describing the whole sequence (pendulum length). We refer to this assessment as \textbf{global} and assume that this attribute does not change over the entire sequence (e.g., user age group). Attributes that change over time are referred to as \textbf{local} (e.g., amount of payment or next purchase category).

\subsection{Sensitivity to permutation}
We randomly permuted order of events along time axis and analyzed performance of pre-trained models. Models were trained without permutation.  The results, presented in Table~\ref{tab:performance_shuffle}, show that the sequence order has only a small effect on most of the results. This suggests that current models may be employing a 'Bag-of-Words' approach, treating the events as a set rather than considering the sequential nature of the data when determining certain \textbf{global} attributes. Additionally, we observed that the dataset's degradation is greater when it has fewer categorical features.

Despite the small degradation, it is still present, highlighting the need for modeling temporal relations.
Furthermore, all the models demonstrate the ability to encode sequential data, as evidenced by a decline in performance on the pendulum dataset where the time component is crucial.

The assessment of the effect of order permutation on the ability to model \textbf{local} attributes is similar to the TPP task, and we leave it for the future work.

\subsection{Sensitivity to data omission}

\begin{figure}
\centering
\begin{subfigure}[t]{0.4\textwidth}
  \centering
  \includegraphics[width=1\textwidth]{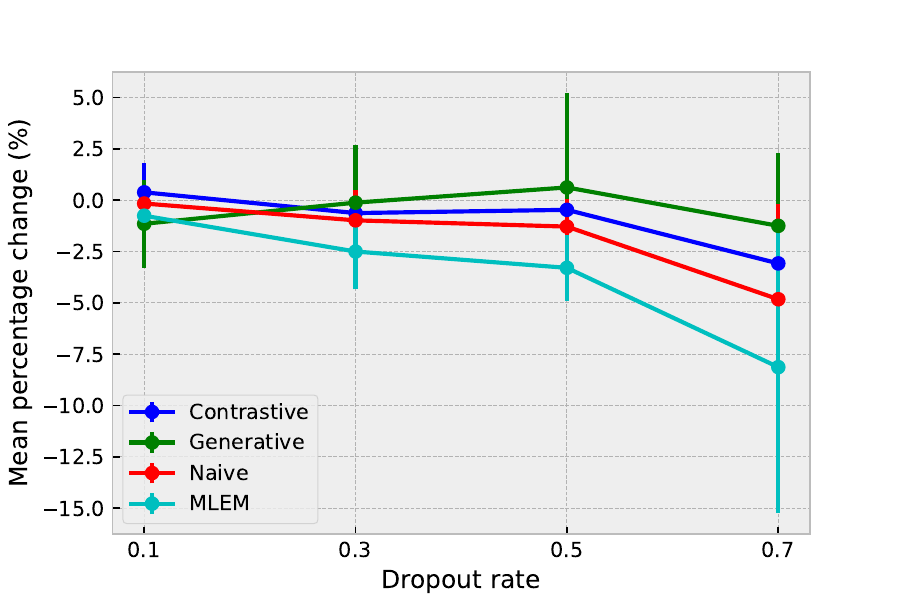}
  \captionsetup{font=small, justification=justified}
  \caption{Mean percentage change averaged across datasets. The X-axis represents the dropout probability, the Y-axis indicates the mean change in the linear probing, compared to the metric with no dropout. Error bars reflect 1 sigma deviation.}
  \label{fig:dropout}
\end{subfigure}%
\hspace{5pt} 
\begin{subfigure}[t]{.55\textwidth}
  \centering
  \includegraphics[width=1\textwidth]{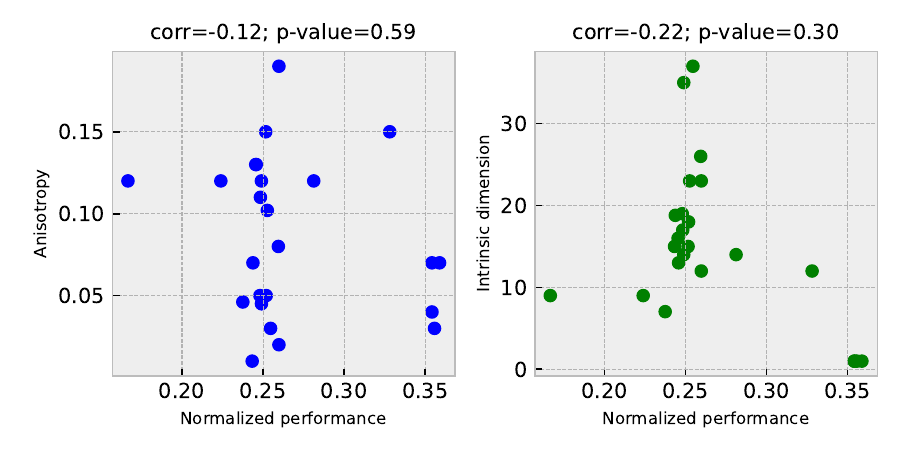}
  \caption{Correlation plots for anisotropy and intrinsic dimension with normalized performance.}
  \label{fig:corr_aid}
\end{subfigure}
\end{figure}

To investigate the effects of data omission, we applied a random dropout strategy, removing entire events from a sequence along the time axis, with varying probabilities $p_{dropout} \in
\{0.1, 0.3, 0.5, 0.7\}$. We report the average percentage decrease in linear probing performance, depicted in Figure~\ref{fig:dropout}.

We notice that MLEM performance slightly decreases at $0.1$ dropout probability and more so at higher levels. This indicates that \textbf{MLEM} is more sensitive to dropout and requires the presence of all the samples rather than depending on a couple specific samples. This property can be used in a sensitivity analysis or related applications. 

Additionally, the generative modeling embeddings exhibited an intriguing behavior: they not only resisted the decline, but actually surpassed their original performance at a dropout rate of 0.5. We believe that such effect show high sensitivity to the presence of noisy or out-of-distribution events. Nevertheless, to draw reliable conclusions, further work is required.



\section{Computational complexity}

To obtain the \textbf{MLEM} encoder, we need to train two models, which is not desirable. However, to demonstrate the practicality of our method, we compare \textbf{MLEM} with SeqNas~\citet{Udovichenko_2024} in terms of performance and computational efficiency. 

\begin{table}[t!]
\caption{Comparison of downstream metrics and computational costs for training \textbf{MLEM} versus \textbf{SeqNas}.}
\centering
\tiny 
\setlength{\tabcolsep}{2pt} 
\resizebox{0.7\textwidth}{!}{
\begin{tabular}{lcccc}
\toprule
Method & \multicolumn{2}{c}{ABank} & \multicolumn{2}{c}{Age} \\
 & ROC-AUC & GPU hours & Accuracy & GPU hours \\
\midrule
MLEM & $0.790 \pm 0.000$ & $47$ & $0.642 \pm 0.005$ & $9$ \\
SeqNas & $0.7963 \pm 0.001$ & $288$ & $0.645 \pm 0.002$ & $80$ \\
\bottomrule
\end{tabular}
}
\label{tab:compute}
\end{table}

SeqNas employs an architectural search to identify the most optimal supervised architecture for the task at hand. We posit that SeqNas represents an upper bound for downstream metrics. Table~\ref{tab:compute} presents results from the two largest datasets, examined by us, demonstrating that \textbf{MLEM} achieves performance near the upper bound set by SeqNas, but with significantly lower computational cost.
Detailed compute budget reported in table~\ref{tab:compute}

\input{content/related_work}

\input{content/discussion}

\input{content/conclusion}

\appendix


\newpage
\bibliographystyle{plainnat} 
\bibliography{main} 

\newpage
\include{content/appendix}

\end{document}

%% file: content/introduction.tex
\section{Introduction}

\textbf{Motivation}. Event Sequences (\ES) are widely used in various real-world applications, including medicine~\citet{waring2020automated}, biology~\citet{valeri2023bioautomated}, social medial analysis~\citet{TMPSurvey}, fault diagnosis~~\citet{fault}, churn prediction~~\citet{churn1,churn2}, customer segmentation~~\citet{segmentation}, fraud detection~~\citet{fraud} and more \citet{zhuzhel2023continuous,fursov2021adversarial}. As a result, there is a growing need to effectively model such data. Most self-supervised methods are focused on images, text, speech, or time-series. In Computer Vision (CV), contrastive learning methods have achieved state-of-the-art results as a pre-training strategy before supervised fine-tuning \citet{he2021masked,chen2020simple,caron2021emerging}. In Natural Language Processing (NLP), most modern models rely on generative pre-training \citet{Radford2018ImprovingLU,devlin2019bert}.  


Self-Supervised Learning (approaches have not been extensively explored for \ES. However, applying existing SSL methods directly to \ES may be unreasonable or even impossible due to several key differences between \ES and other well-studied domains: \textbf{a.  Multiple Features,} events in \ES can be described by multiple features, which adds complexity to the learning process. \textbf{b. Categorical and Numerical Values}, event features in \ES may include both categorical and numerical values, further complicating modeling. \textbf{c. Non-Uniform Temporal Spacing}, events in \ES are not uniformly spaced in time, which requires models to handle irregular temporal patterns. \textbf{d.  Temporal Proximity}, it is to assumed that events occurring in close temporal proximity share relevant information.  Given these unique characteristics of \ES, further research is needed to explore SSL approaches specifically tailored to ES and address these challenges.  We describe a formal structure of \ES in Section~\ref{sec:prelimenaries}.
 
\textbf{The aim of our work} is to answer the following questions: \textbf{Q1:} Which self-supervised approach is superior for ES modeling: generative or contrastive? \textbf{Q2:} Can we efficiently combine generative and contrastive approaches into one procedure to obtain more robust results?




\textbf{The Contributions.} To answer these questions, we \textbf{(1)} study established generative, contrastive, and hybrid approaches. \textbf{(2)}  To the best of our knowledge, we are the first to examine generative pre-training, for \ES classification, and hybrid  methods in the domain of \ES in general. \textbf{(3)} We introduce a new technique named MLEM, inspired by contemporary multimodal studies, such as \citep{radford2021learning,zhai2023sigmoid,viktor2023imad}. MLEM allows for the combination of various SSL approaches by treating differently pre-trained models as distinct modalities. More precisely, MLEM aligns the contrastive and generative embeddings.  We demonstrate that such combination achieves superior performance on average across multiple datasets and metrics compared to existing methods.



    
    
    

\textbf{Evaluation} employs a diverse set of datasets to cover various aspects of the problem space: three real-world transactional datasets from the SeqNAS benchmark \citep{Udovichenko_2024}, the widely-used medical dataset Physionet 2012, and a simulated dataset of pendulum motion designed to examine explicit time dependencies. This selection ensures that our findings are robust across different types of data and tasks. To comprehensively measure performance we employ a set of metrics, including downstream tasks of regression or classification of sequences, prediction of the next event type or timing, intrinsic dimension, and anisotropy.

\textbf{Baselines.} To ensure a clear and fair comparison, we have selected only the top-performing model from each setting for our analysis, comparison of baseline approaches is presented in Table~\ref{approaches_best} and Table~\ref{approaches_ssl}. These include models from recent studies on contrastive learning \citep{babaev2022coles}, generative modeling in \ES\citep{lin2022exploring}, and hybrid approaches from other domains \cite{kim2021hybrid}. We focus on these models because they are clearly superior in their categories, making the inclusion of lesser models unnecessary.

\begin{wrapfigure}{r}{0.4\textwidth} 
  \centering
  \includegraphics[width=0.35\textwidth]{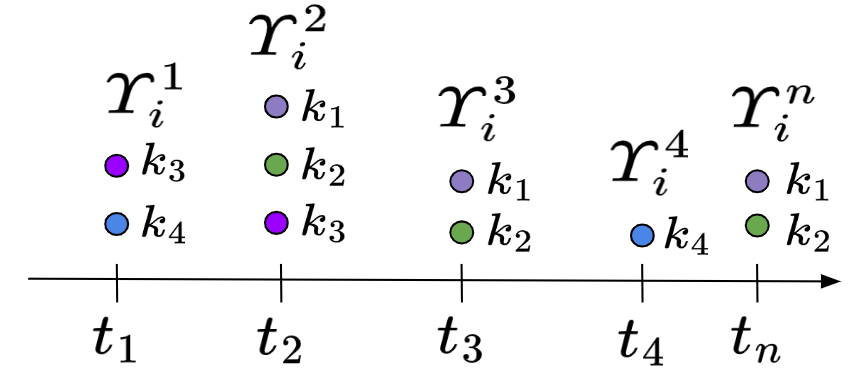}
 
  \caption{An example of one sequence $S_i$ with some categorical and real-valued sub-events $\{k^1,k^2, \ldots\}$.}
  \label{fig:evs_example}
\end{wrapfigure}

We provide the source code for all the experiments described in this paper.\footnote{The source code of our work is publicly available at \url{https://github.com/VityaVitalich/MLEM}.}

%% file: content/preliminary.tex
\section{Event Sequences Preliminaries} \label{sec:prelimenaries}

Datasets used in this work can be described as sets of sequences: $C=\{(S_1,y_1),(S_2,y_2), \ldots\}$, where $y_i$ is an attribute corresponding to the entire sequence or some target. Each $S_i=((t^1_i,\Upsilon^1_i), (t^2_i,\Upsilon^2_i), \ldots)$ is a sequence of events $x^j_i=(t^j_i,\Upsilon^j_i)$, where $t_i^j$ represents the time when the event $x_i^j$ occurred and $t_i^j \leq t^{j+1}_{i}$. The set of sub-events $\Upsilon^j_i=\{k^1,k^2, \ldots\}$ describes the event $x_i^j$. It is important to note that in this work, the terms \textit{Event Sequences} (\ES) and \textit{Irregularly Sampled Time-series} (ISTS) are used interchangeably, as the occurrence of measurement or sampling for the latter can be seen as an event.  Figure~\ref{fig:evs_example} depicts an example of one sequence.


%% file: content/fine_tune_table.tex
\begin{table}[t]
\centering
\begin{minipage}{0.56\textwidth}
\caption{\ES modeling and choice of baselines}
\centering
\resizebox{\textwidth}{!}{%
  \centering
    \begin{tabular}{c|c|c|c}
        & Supervised & Contrastive & Generative \\ \hline
       Paper & SeqNAS & CoLES & Generative TPP \\
       & \citeyear{Udovichenko_2024} & \citeyear{babaev2022coles} & \citeyear{lin2022exploring}  \\  \hline
       Best base models & GRU & GRU & REV-ATT, LSTM 
        \label{approaches_best}
    \end{tabular}
}
\label{approaches_best}
\end{minipage}%
\hspace{3mm} 
\begin{minipage}{0.40\textwidth}
\caption{Hybrid SSL approaches}
\centering
\resizebox{\textwidth}{!}{%
 \begin{tabular}{c|c|c}
        & Sum of Losses (Na\"ive) & Multimodal  \\ \hline
      Paper & DINO~\citeyear{oquab2023dinov2},  &  \\  
      & iBOT~\citeyear{zhou2022ibot} &  MLEM \\ 
      & GCRL~\citeyear{kim2021hybrid} & \\  \hline
      Domain & Images and Text & \ES 
    \end{tabular}
}
\label{approaches_ssl}
\end{minipage}

\vspace{3mm}

\caption{Evaluation of methods fine-tuned with supervised loss for downstream task.  
}
\centering
\resizebox{\textwidth}{!}{
\begin{tabular}{lccccc}
\toprule
Method & \multicolumn{1}{c}{ABank} & \multicolumn{1}{c}{Age} & \multicolumn{1}{c}{PhysioNet} & \multicolumn{1}{c}{Pendulum} & \multicolumn{1}{c}{TaoBao} \\ 
 & {\small ROC-AUC} & {\small Accuracy} & {\small ROC-AUC} & {\small MSE} & {\small ROC-AUC} \\
\hline
Supervised & $0.768 \pm 0.000$ & $0.602 \pm 0.005$ & \underline{$0.790 \pm 0.021$} & $\underline{0.33 \pm 0.02}$ & $0.684 \pm 0.002$ \\ 
\midrule
Contrastive & $0.729 \pm 0.033$ & $0.638 \pm 0.007$ & $\mathbf{0.815 \pm 0.013}$ & $\mathbf{0.26 \pm 0.02}$ & $0.679 \pm 0.003$ \\ 
Generative & $\underline{0.788 \pm 0.003}$ & \underline{$0.639 \pm 0.007$} & $0.787 \pm 0.007$ & $\mathbf{0.26 \pm 0.03}$ & $\mathbf{0.695 \pm 0.004}$ \\ 
\midrule
Na\"ive & $0.658 \pm 0.020$ & $0.638 \pm 0.007$ & $0.759 \pm 0.024$ & $\mathbf{0.26 \pm 0.04}$ & \underline{$0.691 \pm 0.002$} \\ 
MLEM & $\mathbf{0.790 \pm 0.000}$ & $\mathbf{0.642 \pm 0.005}$ & $0.780 \pm 0.004$ & $\mathbf{0.26 \pm 0.05}$ & $\mathbf{0.695 \pm 0.002}$ \\ 
\bottomrule
\end{tabular}
}
\label{tab:fine-tune}
\end{table}

%% file: content/probing_table.tex
\begin{table*}[t!]
\caption{Evaluation of self-supervised methods using linear and non-linear probing for downstream and TPP tasks. 
The best-performing values are highlighted in \textbf{bold}, while the second-best values are \underline{underlined}.}
\setlength{\tabcolsep}{0.1pt}
\centering
\resizebox{\textwidth}{!}{%
    \begin{tabular}{lcccccccccc}
    
    \toprule
    & \multicolumn{2}{c}{ABank} & \multicolumn{2}{c}{Age} & \multicolumn{2}{c}{PhysioNet} & \multicolumn{2}{c}{Pendulum} & \multicolumn{2}{c}{TaoBao} \\
    \cmidrule(lr){2-3} \cmidrule(lr){4-5} \cmidrule(lr){6-7} \cmidrule(lr){8-9} \cmidrule(lr){10-11}
    & ROC-AUC & TPP Acc. & Acc. & TPP Acc. & ROC-AUC  & TPP MSE & MSE & TPP MSE & ROC-AUC & TPP Acc. \\
    \midrule
    & \multicolumn{10}{c}{\textbf{Linear Probing}} \\
    \midrule
    Contrastive & $0.678 \pm 0.003$ & $0.37$ & \underline{$0.623 \pm 0.010$} & $0.28$ & $\mathbf{0.819 \pm 0.003}$ & \underline{$1.05$} & $\mathbf{0.37 \pm 0.02}$ & \underline{$0.8$} & $0.680 \pm 0.001$ & $0.22$ \\
    Generative & \underline{$0.753 \pm 0.003$} & \underline{$0.43$} & $0.610 \pm 0.004$ & \underline{$0.3$} & $0.759 \pm 0.010$ & $\mathbf{0.93}$ & $0.74 \pm 0.09$ & \underline{$0.8$} & $\mathbf{0.689 \pm 0.005}$ & $\mathbf{0.35}$ \\
    Na\"ive & $0.703 \pm 0.002$ & $0.37$ & $0.602 \pm 0.003$ & $\mathbf{0.31}$ & $0.733 \pm 0.005$ & $\mathbf{0.93}$ & $0.59 \pm 0.07$ & \underline{$0.8$} & $0.680 \pm 0.003$ & $0.32$ \\
    MLEM & $\mathbf{0.757 \pm 0.000}$ & $\mathbf{0.43}$ & $\mathbf{0.627 \pm 0.000}$ & $0.29$ & \underline{$0.762 \pm 0.010$} & $\mathbf{0.93}$ & \underline{$0.41 \pm 0.01$} & $\mathbf{0.79}$ & \underline{$0.683 \pm 0.003$} & $\mathbf{0.35}$ \\
    \midrule
    & \multicolumn{10}{c}{\textbf{Non-linear Probing}}  \\
    \midrule
    Contrastive & $0.691 \pm 0.004$ & $0.37$ & \underline{$0.629 \pm 0.010$} & $0.28$ & $\mathbf{0.822 \pm 0.001}$ & \underline{$1.05$} & $\mathbf{0.37 \pm 0.00}$ & \underline{$0.8$} & $0.686 \pm 0.001$ & $0.22$ \\
    Generative & \underline{$0.758 \pm 0.003$} & \underline{$0.43$} & $0.618 \pm 0.004$ & \underline{$0.3$} & $0.764 \pm 0.006$ & $\mathbf{0.93}$ & $0.63 \pm 0.02$ & \underline{$0.8$} & $0.684 \pm 0.002$ & $\mathbf{0.35}$ \\
    Na\"ive & $0.704 \pm 0.005$ & $0.37$ & $0.608 \pm 0.002$ & $\mathbf{0.31}$ & $0.774 \pm 0.008$ & $\mathbf{0.93}$ & $0.57 \pm 0.05$ & \underline{$0.8$} & $\mathbf{0.690 \pm 0.002}$ & $0.32$ \\
    MLEM & $\mathbf{0.759 \pm 0.003}$ & $\mathbf{0.43}$ & $\mathbf{0.634 \pm 0.005}$ & $0.29$ & \underline{$0.780 \pm 0.001$} & $\mathbf{0.93}$ & \underline{$0.40 \pm 0.01$} & $\mathbf{0.79}$ & \underline{$0.688 \pm 0.002$} & $\mathbf{0.35}$ \\
\bottomrule
\end{tabular}
}
\label{tab:linear-non-linear}
\end{table*}

%% file: content/related_work.tex
\section{Related work}
In the following section, we will review various generative, contrastive, hybrid, and other related methods and works.

\textbf{Generative methods} for pre-training and representation learning have seen notable development in recent years in both NLP~\citet{Radford2018ImprovingLU,devlin2019bert,raffel2023exploring} and CV~\citet{he2021masked,assran2023selfsupervised,bachmann2022multimae}. In the \ES domain, the sole study by \citet{lin2022exploring} investigates various generative models for TPP tasks. Based on their findings, we have considered their best performing setting to serve as a generative model for our study, with slight modifications to fit our tasks. 


\textbf{Contrastive methods} are a widely recognized in  \citet{he2020momentum,grill2020bootstrap}.  These methods typically draw closer the embeddings of variations of the same object and distance them from those of different objects.  In CoLES~\citet{babaev2022coles} authors study contrastive learning for \ES. In their work, the positive samples consist of subsequences derived from a single user's events, while the negative samples are obtained from events of different users. We utilize this method as best performing along contrastive approaches. 

\textbf{Hybrid methods.} Our work falls into the category of hybrid self-supervised approaches \citet{qi2023contrast,yang2023generative,lazarow2017introspective} 
Few studies have merged generative and contrastive methods, typically focusing on loss combination. DINOv2 \citet{oquab2023dinov2} updates the traditional contrastive loss from DINO \citet{caron2021emerging} with iBOT's \citet{zhou2022ibot} reconstruction loss through Masked Image Modeling. Similarly, GCRL \citet{kim2021hybrid} combines these losses via a weighted sum, significantly outperforming models that are purely generative or contrastive. We follow the same methodology to form the hybrid baseline method with simply summing contrastive and generative losses. 



 
\textbf{Supervised learning} is commonly used for the \ES classification task. 
The most comprehensive recent paper SeqNAS~\citet{Udovichenko_2024} adopts Neural Architecture Search to identify optimal architectures without strong prior assumptions about the underlying process. A key finding is the effectiveness of Recurrent Neural Networks (RNNs) in modeling \ES, leading us to incorporate RNN encoder in our model. Furthermore, the authors propose benchmark datasets for \ES classification. We selected the largest datasets from the benchmark, as self-supervised methods require substantial data volumes. 

%% file: content/discussion.tex
\section{Limitations}


\textbf{Physionet underperformane}. We observe consistent underperformance of methods that include generative approach on the PhysioNet dataset. This could be attributed to the high rate of missing values, which complicates the generation of accurate predictions for subsequent steps. Furthermore, the relatively small size of the dataset may be a contributing factor, given that generative modeling is generally dependent on large amounts of data.

\textbf{Compute Resources}. One limitation of our model is the requirement for a higher computational budget compared to previous methods. However, in the domain of event sequences, the primary challenge is typically achieving the highest metric rather than optimizing for computational efficiency.

\textbf{Datasets and Models}. We selected the most relevant datasets to cover a broad spectrum of tasks and chose the best-performing models from previous studies to represent SSL methods. Nonetheless, the comparison could be improved by incorporating a wider variety of datasets and model architectures.

\textbf{Scalability}. Although we aimed to ensure a fair comparison, the results may vary with different scales of data or models. However, observed trend show that more data leads to better results on MLEM and generative method.


%% file: content/conclusion.tex
\section{Conclusion}

To our knowledge, our study is the first to apply generative modeling to event sequences for representation learning and pre-training. We found that neither generative nor contrastive pre-training definitively outperforms the other.

In response, we developed the MLEM method, which creatively combines contrastive learning with generative modeling. Inspired by recent advances in multimodal research, MLEM treats these approaches as separate modalities and consistently outperforms existing models on most datasets across both primary and secondary objectives. This finding suggests a promising new direction for enhancing self-supervised learning across various domains.

Additionally, our research provides several key insights. Firstly, the best current self-supervised models do not utilize the sequential nature of data effectively. We also discovered that random event deletion does not adversely affect performance; in some cases, it actually enhances it. Secondly, unlike findings in other domains \citep{dubois2024evaluating,ait-saada-nadif-2023-anisotropy}, we observed no correlation between anisotropy or intrinsic dimension and downstream performance.

%% file: content/appendix.tex
\section*{Contrastive learning} \label{appendix:COLES}

Formally, COLES suggest constructing a set of positive pairs by sampling sub-sequences from a sequence $S_i \rightarrow \{S_i^{'}, S_i^{''}\}$, where each element in $\{S_i^{'}, S_i^{''}\}$ is shorter than $S_i$ and elements in $\{S_i^{'}, S_i^{''}\}$  may intersect (the number of sub-sequences is not limited to two). Those sequences are encoded with RNN and the last hidden state is extracted as latent vector. We refer to COLES sampling as \textbf{subsequence sampler} in Figure~\ref{fig:contrastive}. To learn the encoder it utilizes loss function~(\ref{eq:coles}) from \citep{lecunloss}.



\begin{equation} \label{eq:coles}
\begin{split}
L^{con} & = \frac{1}{|C|} \sum_{i=1}^{|C|} \sum_{j=1}^{|C|} \Bigg( z_{ij}\frac{1}{2}||h_i-h_j||_2^2 + (1-z_{ij})\frac{1}{2}\max\{0, \rho - ||h_i-h_j||_2\}^2 \Bigg)
\end{split}
\end{equation}

Here, \( z_{ij} \in \{0, 1\} \) denotes if objects are semantically similar, \( h \)  is the sequence embedding,  \( \rho \) - the minimal margin between dissimilar objects and $C=\{S_1, S_2, \ldots\}$ is a set of sequences.

\section*{Objectives description} \label{appendix:objectives}

\subsection{Main Objectives} 

To assess the quality of embeddings, we primarily rely on metrics from downstream task. This involves utilizing both linear and non-linear probing methods. For linear probing we simply learn linear or logistic regression on the embeddings. For non-linear probing, we employ Gradient Boosting algorithms with LGBM \citet{NIPS2017_6449f44a} applied directly to the embeddings.

We evaluate embeddings on several tasks, including the prediction of an attribute or a target  $y_i$ given the entire $S_i$ and the prediction of consequent event $x^j$ given a set of $S_i = \{x_i^{1}, \ldots ,x^{j-1}_{i}\}$. The second objective addresses the TPP task, which involves predicting the type of the next event or the time of the next event's arrival. We have processed each dataset such that the target is either the category of the next event (for datasets that include this feature) or the time until the next event (for datasets without a primary event category). 

\subsection{Secondary Objectives}

We assess anisotropy in line with the approach used in \citet{razzhigaev2023shape}. We compute anisotropy as the ratio of the highest singular value to the sum of all singular values:

$$ Anisotropy(H) = \frac{\sigma_1^2}{\sum\limits_{i} \sigma_i^2}$$

To determine the intrinsic dimension, we employ the method proposed in \citet{Facco_2017}. This method examines how the volume of an $n$-dimensional sphere changes with dimension $d$. Further details are available in the original paper or in \citet{razzhigaev2023shape}.

\section*{Detailed Dataset Information} \label{appendix:dataset}
Detailed information regarding all the datasets is presented in Table~\ref{table:datasets}.

\smallskip
\textbf{ABank} is a dataset of bank transactions with regularly sampled time-steps, representing transaction IDs rather than the transaction times. The goal is binary classification: predicting whether a user will default.

\smallskip

\textbf{Age} consists of bank transactions with irregularly sampled time-steps. The task is to categorize users into one of four age groups.

\smallskip

\textbf{PhysioNet} features medical measurements with irregularly sampled time-steps. The target is binary classification: predicting in-hospital mortality.

\smallskip

\textbf{Pendulum} is a synthetic dataset created from pendulum simulations with irregularly sampled time-steps, using the Hawkes process. It includes features like the x and y coordinates of the pendulum, normalized by its length. Each observation has a unique real-valued pendulum length, and the task is to predict this length. The detailed dataset synthesis procedure is described in~section ~\nameref{sec:pendulum}.

\smallskip

\textbf{Taobao} captures user activity on the TaoBao platform with irregular time-steps. The binary classification task is to predict whether a user will make a payment in the next 7 days.

\medskip
For all datasets, categories that appeared fewer than 500 times were consolidated into a single category. Event time values were normalized to a unit interval for consistency. In the case of the PhysioNet dataset, features were aggregated at 360-second intervals. If a feature appeared at least once within this timeframe, its average was calculated. If a feature was absent, it was marked with a value of -1. No other manual feature generation or specific preprocessing was performed, unless otherwise specified.

\smallskip
\textbf{Train - test - validation splits.} For test sets we followed existing literature \citet{babaev2022coles,Udovichenko_2024}, with the exception of the PhysioNet dataset, for which we  created the split. The train and validation sets for each training session were random in fixed proportions. All the metrics are reported on test sets. In cases where sequences exceeded a specified length, we selected the $N$ most recent events, with $N$ being the sequence length as defined in dataset Table~\ref{table:datasets}.

\begin{table*}[ht]
\centering
\caption{Statistics of datasets} 
\resizebox{0.8\textwidth}{!}{
\begin{tabular}{lcccccc}
\toprule
 & ABank  & Age & Taobao & PhysioNet & Pendulum \\
\midrule
\# Observations & 2.7B  & 26B & 7.8M & 0.5M & 8.9M \\
Mean observations per user & 280 & 881 & 806 & 72 & 89 \\
Observations std. per user  & 270  & 124 & 1042 & 21 & 11  \\
Max observations per user in modeling & 200 & 1000 & 1000 & 200 & 200 \\
\# Sequences & 963M  & 30K & 9K & 8K & 100K \\
\# Classes & 2 & 4 & 2 & 2 & - \\
\# Categorical features & 16 & 1 & 2 & 3 & 0 \\
\# Real-valued features & 3 & 1 & 0 & 38 & 2\\
Target & Default & Age group & Payment & Mortality & Length\\
\bottomrule
\end{tabular}
}

\label{table:datasets}
\end{table*}

\section*{Pendulum dataset generation} \label{sec:pendulum}

We developed a pendulum dataset to assess models when time dependency is crucial. 


We simulated pendulum motion with different lengths and sampled coordinates with irregular time intervals, which were derived using a sampling method based on the Hawkes process. Therefore, our dataset consists of sequences where each event is represented with time and two coordinates, each sequence is generated with different pendulum length.  

We opted to Hawkes process to emphasize the critical role of accurate event timing in successful model performance corresponding to real world applications.

\medskip

To model the Hawkes process, we consider the following intensity function $\lambda(t)$ that is given by~(\ref{eq:hawkes}).
\begin{equation}
    \lambda(t) = \mu + \sum_{t_i < t} \alpha e^{-\beta (t - t_i)}
    \label{eq:hawkes}
\end{equation}

We used following parameters for the Hawkes process: 
\begin{itemize}
    \item $\mu$ is the base intensity, was fixed at 10;
    \item $\alpha$ is the excitation factor, was chosen to be 0.2;
    \item $\beta$ is the decay factor, was set to 1.
    \item $t_i$ are the times of previous events before time $t$.
\end{itemize}

\begin{figure}[t] 
  \centering
  \includegraphics[width=\textwidth]{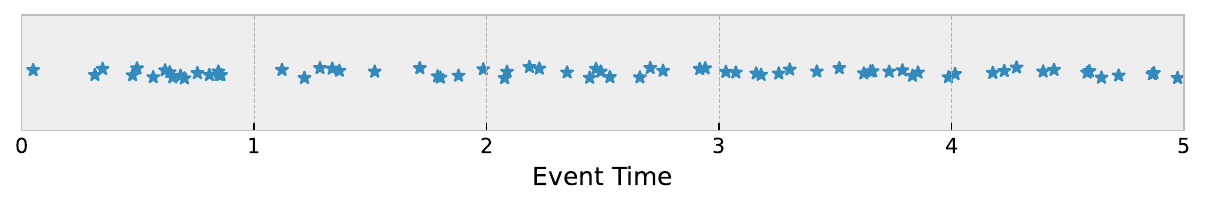}
  \caption{Example of temporal distribution of events ranging from 0 to 5 seconds. Each event is marked with a star along the timeline. The y-axis serves only as a technical aid to separate the events for clarity and does not convey any additional information.}
  \label{fig:hawkes}
\end{figure}

The example of generated event times with these parameters is depicted in Figure~\ref{fig:hawkes}.

\begin{figure}[t] 
  \centering
  \includegraphics[width=\textwidth]{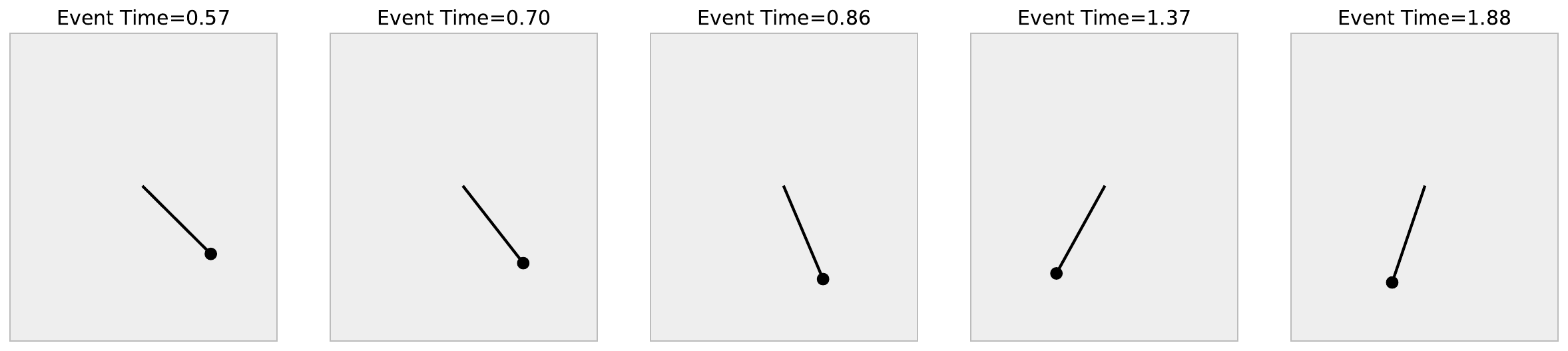}
  \caption{The figure illustrates the pendulum motion at various instances, with time steps determined by a Hawkes process. It captures the pendulum's trajectory using only the normalized planar coordinates at these sampled times.}
  \label{fig:pendulum}
\end{figure}

These parameter values were selected with the intention of generating sequences that contained approximately 100 events each. Additionally, this specific combination of \( \mu \), \( \alpha \), and \( \beta \) was designed to create sequences where events would be densely clustered during certain intervals and sparsely distributed during others. This configuration allowed us to simulate scenarios that closely mimic real-world dynamics, where event occurrences can fluctuate between periods of high and low activity. 

\medskip

To model the pendulum we consider the second-order differential equation:
\begin{equation}
    \theta'' + \left( \frac{b}{m} \right) \theta' + \left( \frac{g}{L} \right) \sin(\theta) = 0
\end{equation}
where,
\begin{itemize}
    \item $\theta''$ is the Angular Acceleration,
    \item $\theta'$ is the Angular Velocity,
    \item $\theta$ is the Angular Displacement,
    \item $b$ is the Damping Factor,
    \item $g = 9.81 \, \text{m/s}^2$ is the acceleration due to gravity,
    \item $L$ is the Length of pendulum,
    \item $m$ is the Mass of bob in kg.
\end{itemize}

To convert this second-order differential equation into two first-order differential equations, we let $\theta_1 = \theta$ and $\theta_2 = \theta'$, which gives us:
\begin{equation}
    \theta_2' = \theta'' = -\left( \frac{b}{m} \right) \theta_2 - \left( \frac{g}{L} \right) \sin(\theta_1)
\end{equation}
\begin{equation}
    \theta_1' = \theta_2
\end{equation}

Thus, the first-order differential equations for the pendulum simulation are:
\begin{align}
    \theta_2' &= -\left( \frac{b}{m} \right) \theta_2 - \left( \frac{g}{L} \right) \sin(\theta_1) \\
    \theta_1' &= \theta_2
\end{align}

\smallskip

In our simulations, we fixed the damping factor $b=0.5$ and the mass of the bob $m=1$. The length $L$ of the pendulum is taken from a uniform distribution $U(0.5, 5)$, representing a range of possible lengths from 0.5 to 5 meters. The initial angular displacement $\theta$ and the initial angular velocity $\theta'$ are both taken from a uniform distribution $U(1, 9)$, which provides a range of initial conditions in radians and radians per second, respectively.

\smallskip

Our primary objective is to predict the length of the pendulum, denoted as $L$, using the normalized coordinates $x$ and $y$ on the plane. These coordinates are scaled with respect to the pendulum's length, such that the trajectory of the pendulum is represented in a unitless fashion. This normalization allows us to abstract the pendulum's motion from its actual physical dimensions and instead focus on the pattern of movement. An illustrative example of this motion is presented in Figure~\ref{fig:pendulum}, where the path traced by the pendulum bob is depicted over time.

\section*{Generated datasets}

\begin{figure}[t] 
  \centering
  \includegraphics[width=0.8\textwidth]{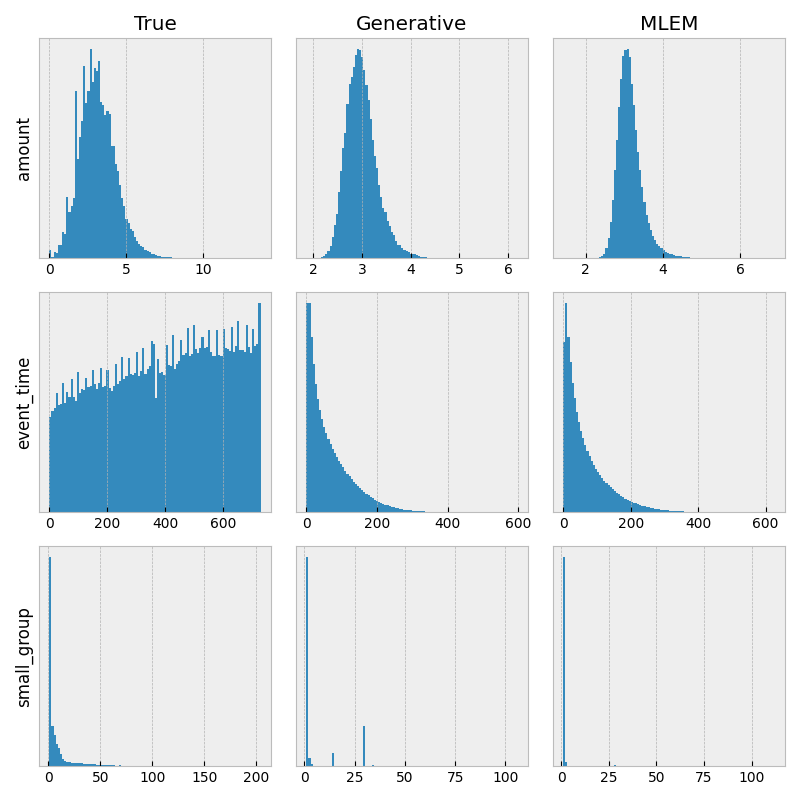}
  \caption{Distributions of features for real and generated Age dataset. True denotes distributions for real dataset, Generative and MLEM denote distributions for datasets generated from embeddigs obtain via Generative and MLEM approaches.}
  \label{fig:gens}
\end{figure}

We conducted a preliminary evaluation of our decoder's ability to reconstruct sequences from their embeddings. We hypothesize that while the regenerated sequences may exhibit slight deviations from the original data, the overall distribution of features across whole dataset should align. To investigate this, we compared distributions of features in  generated datasets, results are visualized in Figure~\ref{fig:gens} for Age dataset.

\medskip

There is a notable resemblance between the generated sequences and the actual data regarding the distribution of numerical feature ''amount'', particularly around the mean. However, the model struggles with accurately reproducing the timing and the key dataset feature—the user group. The MLEM tends to overrepresent the most frequent classes while underrepresenting less common ones. Moreover, the Generative model despite of the mostly same performance exhibits unexpected behaviour, overproducing some of the rarer classes.

\smallskip

These findings suggest directions for potential improvements, more specifically: improving time component modeling, applying different generation approaches, and studying different architecture designs such as enhancing either the encoder's or decoder's performance.

\begin{table}[t]
\centering
\caption{GPU Hours spent on full model training with single Nvidia A100}
\begin{tabular}{lccccc}
\hline
\textbf{Method} & \textbf{ABank} & \textbf{Age} & \textbf{PhysioNet} & \textbf{TaoBao} & \textbf{Pendulum} \\
\hline
Supervised       & 7        & 1        & $<$ 1    & $<$ 1   & $<$ 1   \\
Contrastive & 28       & 2        & 1        & 1       & 3       \\
Generative         & 19       & 7        & 1        & 1       & 2       \\
Na\"ive       & 45       & 6        & 2        & 2       & 8       \\
MLEM        & 47  & 9    & 2    & 2   & 5   \\
\hline
\end{tabular}
\label{tab:compute}
\end{table}

%% file: main.bbl
\begin{thebibliography}{46}
\providecommand{\natexlab}[1]{#1}
\providecommand{\url}[1]{\texttt{#1}}
\expandafter\ifx\csname urlstyle\endcsname\relax
  \providecommand{\doi}[1]{doi: #1}\else
  \providecommand{\doi}{doi: \begingroup \urlstyle{rm}\Url}\fi

\bibitem[Ait-Saada and Nadif(2023)]{ait-saada-nadif-2023-anisotropy}
Mira Ait-Saada and Mohamed Nadif.
\newblock Is anisotropy truly harmful? a case study on text clustering.
\newblock In Anna Rogers, Jordan Boyd-Graber, and Naoaki Okazaki, editors, \emph{Proceedings of the 61st Annual Meeting of the Association for Computational Linguistics (Volume 2: Short Papers)}, pages 1194--1203, Toronto, Canada, July 2023. Association for Computational Linguistics.
\newblock \doi{10.18653/v1/2023.acl-short.103}.
\newblock URL \url{https://aclanthology.org/2023.acl-short.103}.

\bibitem[Assran et~al.(2023)Assran, Duval, Misra, Bojanowski, Vincent, Rabbat, LeCun, and Ballas]{assran2023selfsupervised}
Mahmoud Assran, Quentin Duval, Ishan Misra, Piotr Bojanowski, Pascal Vincent, Michael Rabbat, Yann LeCun, and Nicolas Ballas.
\newblock Self-supervised learning from images with a joint-embedding predictive architecture, 2023.

\bibitem[Babaev et~al.(2022)Babaev, Ovsov, Kireev, Ivanova, Gusev, Nazarov, and Tuzhilin]{babaev2022coles}
Dmitrii Babaev, Nikita Ovsov, Ivan Kireev, Maria Ivanova, Gleb Gusev, Ivan Nazarov, and Alexander Tuzhilin.
\newblock Coles: contrastive learning for event sequences with self-supervision.
\newblock In \emph{Proceedings of the 2022 International Conference on Management of Data}, pages 1190--1199, 2022.

\bibitem[Bachmann et~al.(2022)Bachmann, Mizrahi, Atanov, and Zamir]{bachmann2022multimae}
Roman Bachmann, David Mizrahi, Andrei Atanov, and Amir Zamir.
\newblock Multimae: Multi-modal multi-task masked autoencoders, 2022.

\bibitem[Carnein and Trautmann(2019)]{segmentation}
Matthias Carnein and Heike Trautmann.
\newblock Customer segmentation based on transactional data using stream clustering.
\newblock In \emph{Advances in Knowledge Discovery and Data Mining: 23rd Pacific-Asia Conference, PAKDD 2019, Macau, China, April 14-17, 2019, Proceedings, Part I 23}, pages 280--292. Springer, 2019.

\bibitem[Caron et~al.(2021)Caron, Touvron, Misra, Jégou, Mairal, Bojanowski, and Joulin]{caron2021emerging}
Mathilde Caron, Hugo Touvron, Ishan Misra, Hervé Jégou, Julien Mairal, Piotr Bojanowski, and Armand Joulin.
\newblock Emerging properties in self-supervised vision transformers, 2021.

\bibitem[Chen et~al.(2020)Chen, Kornblith, Norouzi, and Hinton]{chen2020simple}
Ting Chen, Simon Kornblith, Mohammad Norouzi, and Geoffrey Hinton.
\newblock A simple framework for contrastive learning of visual representations, 2020.

\bibitem[Devlin et~al.(2019)Devlin, Chang, Lee, and Toutanova]{devlin2019bert}
Jacob Devlin, Ming-Wei Chang, Kenton Lee, and Kristina Toutanova.
\newblock Bert: Pre-training of deep bidirectional transformers for language understanding, 2019.

\bibitem[Dubois et~al.(2021)Dubois, Kiela, Schwab, and Vedantam]{dubois2021learning}
Yann Dubois, Douwe Kiela, David~J. Schwab, and Ramakrishna Vedantam.
\newblock Learning optimal representations with the decodable information bottleneck, 2021.

\bibitem[Dubois et~al.(2024)Dubois, Hashimoto, and Liang]{dubois2024evaluating}
Yann Dubois, Tatsunori Hashimoto, and Percy Liang.
\newblock Evaluating self-supervised learning via risk decomposition, 2024.

\bibitem[Facco et~al.(2017)Facco, d’Errico, Rodriguez, and Laio]{Facco_2017}
Elena Facco, Maria d’Errico, Alex Rodriguez, and Alessandro Laio.
\newblock Estimating the intrinsic dimension of datasets by a minimal neighborhood information.
\newblock \emph{Scientific Reports}, 7\penalty0 (1), September 2017.
\newblock ISSN 2045-2322.
\newblock \doi{10.1038/s41598-017-11873-y}.
\newblock URL \url{http://dx.doi.org/10.1038/s41598-017-11873-y}.

\bibitem[Fursov et~al.(2021)Fursov, Morozov, Kaploukhaya, Kovtun, Rivera-Castro, Gusev, Babaev, Kireev, Zaytsev, and Burnaev]{fursov2021adversarial}
Ivan Fursov, Matvey Morozov, Nina Kaploukhaya, Elizaveta Kovtun, Rodrigo Rivera-Castro, Gleb Gusev, Dmitry Babaev, Ivan Kireev, Alexey Zaytsev, and Evgeny Burnaev.
\newblock Adversarial attacks on deep models for financial transaction records, 2021.

\bibitem[Grill et~al.(2020)Grill, Strub, Altché, Tallec, Richemond, Buchatskaya, Doersch, Pires, Guo, Azar, Piot, Kavukcuoglu, Munos, and Valko]{grill2020bootstrap}
Jean-Bastien Grill, Florian Strub, Florent Altché, Corentin Tallec, Pierre~H. Richemond, Elena Buchatskaya, Carl Doersch, Bernardo~Avila Pires, Zhaohan~Daniel Guo, Mohammad~Gheshlaghi Azar, Bilal Piot, Koray Kavukcuoglu, Rémi Munos, and Michal Valko.
\newblock Bootstrap your own latent: A new approach to self-supervised learning, 2020.

\bibitem[Hadsell et~al.(2006)Hadsell, Chopra, and Lecun]{lecunloss}
Raia Hadsell, Sumit Chopra, and Yann Lecun.
\newblock Dimensionality reduction by learning an invariant mapping.
\newblock pages 1735 -- 1742, 02 2006.
\newblock ISBN 0-7695-2597-0.
\newblock \doi{10.1109/CVPR.2006.100}.

\bibitem[He et~al.(2020)He, Fan, Wu, Xie, and Girshick]{he2020momentum}
Kaiming He, Haoqi Fan, Yuxin Wu, Saining Xie, and Ross Girshick.
\newblock Momentum contrast for unsupervised visual representation learning, 2020.

\bibitem[He et~al.(2021)He, Chen, Xie, Li, Dollár, and Girshick]{he2021masked}
Kaiming He, Xinlei Chen, Saining Xie, Yanghao Li, Piotr Dollár, and Ross Girshick.
\newblock Masked autoencoders are scalable vision learners, 2021.

\bibitem[Jain et~al.(2021)Jain, Khunteta, and Srivastava]{churn1}
Hemlata Jain, Ajay Khunteta, and Sumit Srivastava.
\newblock Telecom churn prediction and used techniques, datasets and performance measures: a review.
\newblock \emph{Telecommunication Systems}, 76:\penalty0 613--630, 2021.

\bibitem[Ke et~al.(2017)Ke, Meng, Finley, Wang, Chen, Ma, Ye, and Liu]{NIPS2017_6449f44a}
Guolin Ke, Qi~Meng, Thomas Finley, Taifeng Wang, Wei Chen, Weidong Ma, Qiwei Ye, and Tie-Yan Liu.
\newblock Lightgbm: A highly efficient gradient boosting decision tree.
\newblock In I.~Guyon, U.~Von Luxburg, S.~Bengio, H.~Wallach, R.~Fergus, S.~Vishwanathan, and R.~Garnett, editors, \emph{Advances in Neural Information Processing Systems}, volume~30. Curran Associates, Inc., 2017.
\newblock URL \url{https://proceedings.neurips.cc/paper_files/paper/2017/file/6449f44a102fde848669bdd9eb6b76fa-Paper.pdf}.

\bibitem[Kim et~al.(2021)Kim, Kim, and Lee]{kim2021hybrid}
Saehoon Kim, Sungwoong Kim, and Juho Lee.
\newblock Hybrid generative-contrastive representation learning, 2021.

\bibitem[Lazarow et~al.(2017)Lazarow, Jin, and Tu]{lazarow2017introspective}
Justin Lazarow, Long Jin, and Zhuowen Tu.
\newblock Introspective neural networks for generative modeling.
\newblock In \emph{Proceedings of the IEEE International Conference on Computer Vision}, pages 2774--2783, 2017.

\bibitem[Liguori et~al.(2023)Liguori, Caroprese, Minici, Veloso, Spinnato, Nanni, Manco, and Gama]{TMPSurvey}
Angelica Liguori, Luciano Caroprese, Marco Minici, Bruno Veloso, Francesco Spinnato, Mirco Nanni, Giuseppe Manco, and Joao Gama.
\newblock Modeling events and interactions through temporal processes--a survey.
\newblock \emph{arXiv preprint arXiv:2303.06067}, 2023.

\bibitem[Lin et~al.(2021)Lin, Tan, Wu, Gao, Li, et~al.]{lin2021empirical}
Haitao Lin, Cheng Tan, Lirong Wu, Zhangyang Gao, Stan Li, et~al.
\newblock An empirical study: Extensive deep temporal point process.
\newblock \emph{arXiv preprint arXiv:2110.09823}, 2021.

\bibitem[Lin et~al.(2022)Lin, Wu, Zhao, Liu, and Li]{lin2022exploring}
Haitao Lin, Lirong Wu, Guojiang Zhao, Pai Liu, and Stan~Z Li.
\newblock Exploring generative neural temporal point process.
\newblock \emph{arXiv preprint arXiv:2208.01874}, 2022.

\bibitem[Moskvoretskii and Kuznetsov(2023)]{viktor2023imad}
Viktor Moskvoretskii and Denis Kuznetsov.
\newblock Imad: Image-augmented multi-modal dialogue.
\newblock \emph{arXiv preprint arXiv:2305.10512}, 2023.

\bibitem[Nakada and Imaizumi(2020)]{nakada2020adaptive}
Ryumei Nakada and Masaaki Imaizumi.
\newblock Adaptive approximation and generalization of deep neural network with intrinsic dimensionality, 2020.

\bibitem[Oquab et~al.(2023)Oquab, Darcet, Moutakanni, Vo, Szafraniec, Khalidov, Fernandez, Haziza, Massa, El-Nouby, Assran, Ballas, Galuba, Howes, Huang, Li, Misra, Rabbat, Sharma, Synnaeve, Xu, Jegou, Mairal, Labatut, Joulin, and Bojanowski]{oquab2023dinov2}
Maxime Oquab, Timothée Darcet, Théo Moutakanni, Huy Vo, Marc Szafraniec, Vasil Khalidov, Pierre Fernandez, Daniel Haziza, Francisco Massa, Alaaeldin El-Nouby, Mahmoud Assran, Nicolas Ballas, Wojciech Galuba, Russell Howes, Po-Yao Huang, Shang-Wen Li, Ishan Misra, Michael Rabbat, Vasu Sharma, Gabriel Synnaeve, Hu~Xu, Hervé Jegou, Julien Mairal, Patrick Labatut, Armand Joulin, and Piotr Bojanowski.
\newblock Dinov2: Learning robust visual features without supervision, 2023.

\bibitem[Qi et~al.(2023)Qi, Dong, Fan, Ge, Zhang, Ma, and Yi]{qi2023contrast}
Zekun Qi, Runpei Dong, Guofan Fan, Zheng Ge, Xiangyu Zhang, Kaisheng Ma, and Li~Yi.
\newblock Contrast with reconstruct: Contrastive 3d representation learning guided by generative pretraining.
\newblock \emph{arXiv preprint arXiv:2302.02318}, 2023.

\bibitem[Radford and Narasimhan(2018)]{Radford2018ImprovingLU}
Alec Radford and Karthik Narasimhan.
\newblock Improving language understanding by generative pre-training.
\newblock 2018.
\newblock URL \url{https://api.semanticscholar.org/CorpusID:49313245}.

\bibitem[Radford et~al.(2021)Radford, Kim, Hallacy, Ramesh, Goh, Agarwal, Sastry, Askell, Mishkin, Clark, Krueger, and Sutskever]{radford2021learning}
Alec Radford, Jong~Wook Kim, Chris Hallacy, Aditya Ramesh, Gabriel Goh, Sandhini Agarwal, Girish Sastry, Amanda Askell, Pamela Mishkin, Jack Clark, Gretchen Krueger, and Ilya Sutskever.
\newblock Learning transferable visual models from natural language supervision, 2021.

\bibitem[Raffel et~al.(2023)Raffel, Shazeer, Roberts, Lee, Narang, Matena, Zhou, Li, and Liu]{raffel2023exploring}
Colin Raffel, Noam Shazeer, Adam Roberts, Katherine Lee, Sharan Narang, Michael Matena, Yanqi Zhou, Wei Li, and Peter~J. Liu.
\newblock Exploring the limits of transfer learning with a unified text-to-text transformer, 2023.

\bibitem[Razzhigaev et~al.(2023)Razzhigaev, Mikhalchuk, Goncharova, Oseledets, Dimitrov, and Kuznetsov]{razzhigaev2023shape}
Anton Razzhigaev, Matvey Mikhalchuk, Elizaveta Goncharova, Ivan Oseledets, Denis Dimitrov, and Andrey Kuznetsov.
\newblock The shape of learning: Anisotropy and intrinsic dimensions in transformer-based models, 2023.

\bibitem[Rubanova et~al.(2019)Rubanova, Chen, and Duvenaud]{rubanova2019latent}
Yulia Rubanova, Ricky T.~Q. Chen, and David Duvenaud.
\newblock Latent odes for irregularly-sampled time series, 2019.

\bibitem[Shao et~al.(2019)Shao, Yan, Lu, Wang, and Gao]{fault}
Siyu Shao, Ruqiang Yan, Yadong Lu, Peng Wang, and Robert~X Gao.
\newblock Dcnn-based multi-signal induction motor fault diagnosis.
\newblock \emph{IEEE Transactions on Instrumentation and Measurement}, 69\penalty0 (6):\penalty0 2658--2669, 2019.

\bibitem[Silva et~al.(2012)Silva, Moody, Scott, Celi, and Mark]{physionet}
Ikaro Silva, George Moody, Daniel~J Scott, Leo~A Celi, and Roger~G Mark.
\newblock Predicting in-hospital mortality of icu patients: The physionet/computing in cardiology challenge 2012.
\newblock In \emph{2012 Computing in Cardiology}, pages 245--248. IEEE, 2012.

\bibitem[Sudharsan and Ganesh(2022)]{churn2}
R~Sudharsan and EN~Ganesh.
\newblock A swish rnn based customer churn prediction for the telecom industry with a novel feature selection strategy.
\newblock \emph{Connection Science}, 34\penalty0 (1):\penalty0 1855--1876, 2022.

\bibitem[Tonekaboni et~al.(2021)Tonekaboni, Eytan, and Goldenberg]{tonekaboni2021unsupervised}
Sana Tonekaboni, Danny Eytan, and Anna Goldenberg.
\newblock Unsupervised representation learning for time series with temporal neighborhood coding, 2021.

\bibitem[Udovichenko et~al.(2024)Udovichenko, Shvetsov, Divitsky, Osin, Trofimov, Sukharev, Glushenko, Berestnev, and Burnaev]{Udovichenko_2024}
Igor Udovichenko, Egor Shvetsov, Denis Divitsky, Dmitry Osin, Ilya Trofimov, Ivan Sukharev, Anatoliy Glushenko, Dmitry Berestnev, and Evgeny Burnaev.
\newblock Seq{NAS}: Neural architecture search for event sequence classification.
\newblock \emph{IEEE Access}, 12:\penalty0 3898–3909, 2024.
\newblock ISSN 2169-3536.
\newblock \doi{10.1109/access.2024.3349497}.
\newblock URL \url{http://dx.doi.org/10.1109/ACCESS.2024.3349497}.

\bibitem[Valeri et~al.(2023)Valeri, Soenksen, Collins, Ramesh, Cai, Powers, Angenent-Mari, Camacho, Wong, Lu, et~al.]{valeri2023bioautomated}
Jacqueline~A Valeri, Luis~R Soenksen, Katherine~M Collins, Pradeep Ramesh, George Cai, Rani Powers, Nicolaas~M Angenent-Mari, Diogo~M Camacho, Felix Wong, Timothy~K Lu, et~al.
\newblock Bioautomated: An end-to-end automated machine learning tool for explanation and design of biological sequences.
\newblock \emph{Cell Systems}, 14\penalty0 (6):\penalty0 525--542, 2023.

\bibitem[Vaswani et~al.(2017)Vaswani, Shazeer, Parmar, Uszkoreit, Jones, Gomez, Kaiser, and Polosukhin]{vaswani2023attention}
Ashish Vaswani, Noam Shazeer, Niki Parmar, Jakob Uszkoreit, Llion Jones, Aidan~N Gomez, {\L}ukasz Kaiser, and Illia Polosukhin.
\newblock Attention is all you need.
\newblock \emph{Advances in neural information processing systems}, 30, 2017.

\bibitem[Waring et~al.(2020)Waring, Lindvall, and Umeton]{waring2020automated}
Jonathan Waring, Charlotta Lindvall, and Renato Umeton.
\newblock Automated machine learning: Review of the state-of-the-art and opportunities for healthcare.
\newblock \emph{Artificial intelligence in medicine}, 104:\penalty0 101822, 2020.

\bibitem[Xie et~al.(2022)Xie, Liu, Yan, Jiang, and Zhou]{fraud}
Yu~Xie, Guanjun Liu, Chungang Yan, Changjun Jiang, and MengChu Zhou.
\newblock Time-aware attention-based gated network for credit card fraud detection by extracting transactional behaviors.
\newblock \emph{IEEE Transactions on Computational Social Systems}, 2022.

\bibitem[Yang et~al.(2023)Yang, Wu, Wu, Zhang, Hong, Zhang, Zhou, and Wang]{yang2023generative}
Yonghui Yang, Zhengwei Wu, Le~Wu, Kun Zhang, Richang Hong, Zhiqiang Zhang, Jun Zhou, and Meng Wang.
\newblock Generative-contrastive graph learning for recommendation.
\newblock 2023.

\bibitem[Yoon et~al.(2019)Yoon, Jarrett, and van~der Schaar]{NEURIPS2019_c9efe5f2}
Jinsung Yoon, Daniel Jarrett, and Mihaela van~der Schaar.
\newblock Time-series generative adversarial networks.
\newblock In H.~Wallach, H.~Larochelle, A.~Beygelzimer, F.~d\textquotesingle Alch\'{e}-Buc, E.~Fox, and R.~Garnett, editors, \emph{Advances in Neural Information Processing Systems}, volume~32. Curran Associates, Inc., 2019.
\newblock URL \url{https://proceedings.neurips.cc/paper_files/paper/2019/file/c9efe5f26cd17ba6216bbe2a7d26d490-Paper.pdf}.

\bibitem[Zhai et~al.(2023)Zhai, Mustafa, Kolesnikov, and Beyer]{zhai2023sigmoid}
Xiaohua Zhai, Basil Mustafa, Alexander Kolesnikov, and Lucas Beyer.
\newblock Sigmoid loss for language image pre-training, 2023.

\bibitem[Zhou et~al.(2022)Zhou, Wei, Wang, Shen, Xie, Yuille, and Kong]{zhou2022ibot}
Jinghao Zhou, Chen Wei, Huiyu Wang, Wei Shen, Cihang Xie, Alan Yuille, and Tao Kong.
\newblock ibot: Image bert pre-training with online tokenizer, 2022.

\bibitem[Zhuzhel et~al.(2023)Zhuzhel, Grabar, Boeva, Zabolotnyi, Stepikin, Zholobov, Ivanova, Orlov, Kireev, Burnaev, et~al.]{zhuzhel2023continuous}
Vladislav Zhuzhel, Vsevolod Grabar, Galina Boeva, Artem Zabolotnyi, Alexander Stepikin, Vladimir Zholobov, Maria Ivanova, Mikhail Orlov, Ivan Kireev, Evgeny Burnaev, et~al.
\newblock Continuous-time convolutions model of event sequences.
\newblock \emph{arXiv preprint arXiv:2302.06247}, 2023.

\end{thebibliography}
